\documentclass[conference]{IEEEtran}
\IEEEoverridecommandlockouts
\usepackage{array}
\usepackage{cite}
\usepackage{amsmath,amssymb,amsfonts}
\usepackage{algorithmic}
\usepackage{graphicx}
\usepackage{textcomp}
\usepackage{xcolor}
\usepackage{url}
\def\BibTeX{{\rm B\kern-.05em{\sc i\kern-.025em b}\kern-.08em
    T\kern-.1667em\lower.7ex\hbox{E}\kern-.125emX}}

\makeatletter
\def\ps@IEEEtitlepagestyle{%
  \def\@oddhead{\hfill
    \parbox[t]{\dimexpr\textwidth-2cm}{\raggedleft
      \footnotesize \textit{Proc. of the 11th International Conference on Engineering and Emerging Technologies (ICEET)\\
      22-23 October 2025, Kuala Lumpur, Malaysia}}}%
  \def\@oddfoot{\parbox[b]{\textwidth}{\vspace*{1em}%
      \footnotesize 979-8-3315-6755-2/25/\$31.00 \textcopyright2025 IEEE}}%
}
\makeatother

\begin{document}

\title{THETA: Triangulated Hand-State Estimation for Teleoperation and Automation in 
Robotic Hand Control\\
{\footnotesize \textsuperscript{}}
}

\author{\IEEEauthorblockN{Akshay Karthik}
\IEEEauthorblockA{
\textit{Arizona State University}\\
Tempe, AZ, USA \\
akshay.karthik08@gmail.com}
\and
\IEEEauthorblockN{Alex Huang}
\IEEEauthorblockA{\textit{Arizona State University} \\
Tempe, AZ, USA \\
alexahuanga1029@gmail.com}
}

\maketitle

\begin{abstract}
The teleoperation of robotic hands is limited by the high costs of depth cameras and sensor gloves, commonly used to estimate hand relative joint positions (XYZ). We present a novel, cost-effective approach using three webcams for triangulation-based tracking to approximate relative joint angles (theta) of human fingers. We also introduce a modified DexHand, a low-cost robotic hand from TheRobotStudio, to demonstrate THETA’s real-time application. Data collection involved 40 distinct hand gestures using three 640x480p webcams arranged at 120-degree intervals, generating over 48,000 RGB images. Joint angles were manually determined by measuring midpoints of the MCP, PIP, and DIP finger joints. Captured RGB frames were processed using a DeepLabV3 segmentation model with a ResNet-50 backbone for multi-scale hand segmentation. The segmented images were then HSV-filtered and fed into THETA’s architecture, consisting of a MobileNetV2-based CNN classifier optimized for hierarchical spatial feature extraction and a 9-channel input tensor encoding multi-perspective hand representations. The classification model maps segmented hand views into discrete joint angles, achieving 97.18\% accuracy, 98.72\% recall, F1 Score of 0.9274, and a precision of 0.8906. In real-time inference, THETA captures simultaneous frames, segments hand regions, filters them, and compiles a 9-channel tensor for classification. Joint-angle predictions are relayed via serial to an Arduino, enabling the DexHand to replicate hand movements. Future research will increase dataset diversity, integrate wrist tracking, and apply computer vision techniques such as OpenAI-Vision. THETA potentially ensures cost-effective, user-friendly teleoperation for medical, linguistic, and manufacturing applications.
\end{abstract}


\section{Introduction}
The robotic teleoperation field, where robotic systems mimic human movements, has been rapidly growing over recent years. This growth is supported by increasing demands for automation in environments that are hazardous, inaccessible, or require high precision. Teleoperated systems are playing key roles in various applications such as surgical robotics, repair industry, nuclear decommissioning, space missions, care for the elderly, and supply chain logistics. A core element of such applications is the robotic hand, which must emulate fine human motor control with high compliance and precision. 

From remotely suturing in telesurgery to managing mechanical parts in nuclear reactors, robotic hands are increasingly being used for operations once only feasible by humans. This expansion is reflected economically: the teleoperated robotics market is projected to grow from \$40.17 billion in 2022 to over \$170 billion by 2032, highlighting the urgency for scalable, adaptable, and cost-effective solutions [1]. At the heart of effective robotic hand teleoperation is the ability to accurately estimate human finger joint angles in real-time, allowing the robotic counterpart to mimic complex hand movements. Despite recent technological breakthroughs, current solutions fall short in balancing accuracy, cost, and accessibility, especially in unstructured, real-world settings. 

In order to support the growing applications of teleoperated robotic hands, robotic hand systems must not only reproduce static hand postures but also reproduce dynamic, real-time human movement. This requires the continuous capture and interpretation of fine-grained biomechanical data, particularly joint angles of the fingers and hand, which are required in a form that is responsive, low-latency, and insensitive to environmental variation. From cutting precise sutures within a medical operating theater to maneuvering tools within a hazardous manufacturing setting, robotic hands must achieve a great degree of articulation in order to serve as reliable proxies for human dexterity. Therefore, the effectiveness of teleoperated robot hands increasingly hinges upon the efficiency and accuracy of supporting motion tracking and hand state estimation algorithms.

\section{Related Work}

\subsection{\textit{Depth-Sensing \& Infrared Systems} }

Depth-sensing and infrared (IR) cameras such as the Intel RealSense D455 and Microsoft Azure Kinect produce 3D point clouds through stereo vision or time-of-flight methods, enabling joint localization. However, their accuracy degrades under occlusion, non-frontal orientations, and motion blur. Gallo et al. illustrated how depth readings start to get noisy and unreliable as soon as the surface normals of the hand are displaced out of the camera's line of sight [2].

High-end IR motion capture systems, such as Vicon, achieve high spatial precision by tracking reflective markers, widely applied in biomechanics and animation. These systems, however, require extensive infrastructure, strict calibration, and significant financial investment, often exceeding \$10,000, which limits their practical use [3]. Additionally, they are sensitive to ambient lighting and require controlled environments [4].

Both classes of systems calculate joint angles by forming vectors between adjacent joints and applying the cosine law. The angle at a joint is computed by the inverse cosine of the normalized dot product between vectors connecting neighboring segments. Although mathematically straightforward, this process depends entirely on consistent and high-quality 3D tracking, which remains a major barrier for depth-based and marker-based systems in unconstrained environments.

 \subsection{\textit{Sensor Glove-Based Systems} }

Sensor gloves, including the CyberGlove II and Manus Prime X, integrate flex sensors and inertial measurement units (IMUs) to capture joint angles and hand orientation in real time, supporting applications in rehabilitation, virtual reality, and robotics. Despite high temporal resolution, these devices face persistent limitations. Dipietro et al. identified calibration drift, mechanical fatigue, and user discomfort during extended use as key concerns [5]. Further, sensor gloves require precise fitting, restrict natural hand movements, and remain costly, with prices exceeding \$10,000 for CyberGlove II and \$5,000 for Manus Prime X, limiting accessibility for general users [6]. 

  \subsection{\textit{Vision-Based Estimation with RGB Cameras} }

Vision-based methods utilizing standard RGB cameras and machine learning have emerged as cost-effective alternatives for hand pose estimation. These approaches employ convolutional neural networks (CNNs) to detect hand landmarks or segment hand regions for joint angle prediction from 2D imagery.

Google’s MediaPipe, for instance, estimates 21 hand landmarks and computes joint angles using geometric calculations. While computationally efficient, its accuracy diminishes under hand rotations, occlusions, or non-frontal poses [7].

More robust methods apply semantic segmentation networks such as U-Net, Mask R-CNN, SegNet, PSPNet, Attention U-Net, Cascade Mask R-CNN, and dilated convolution models to isolate hand regions prior to pose estimation, improving robustness to background noise and lighting variations [8]. However, most of these models rely on single-view 2D inputs and cannot resolve depth ambiguity under occlusion. Their limited perspective restricts their ability to capture full 3D hand motion, highlighting the need for multi-view or triangulation-based approaches that can reconstruct hand poses from multiple viewpoints for reliable joint angle estimation.

\section{Methodology}
This paper presents a novel, low-cost pipeline for real-time finger joint angle estimation from a multi-view vision system and deep learning for robotic hand teleoperation. The pipeline has four major steps: hardware integration and setup, dataset annotation and creation, segmentation and classification, and real-time actuation of a robotic hand. Each of these steps was designed to be robust, low-cost, and accurate.

The process begins with the construction of a 3D-printed, servo-driven robotic hand and the development of a ROS 2-based control system for actuation. Next, synchronized multi-angle webcam images of hand gestures are collected, segmented using DeepLabV3, and manually annotated with corresponding joint angles. These segmented images are then fed into a MobileNetV2-based classifier trained to predict joint angles efficiently and accurately. Finally, the predicted joint angles are transmitted in real time to control the robotic hand, completing the teleoperation loop with live inference and actuation.

\subsection{DexHand Robotic Hand Design \& ROS 2 Control}\label{AA}
The hardware for the robotic hand used in the project was derived from TheRobotStudio's open-source DexHand V1.0 design [9]. The design was remodeled using a combination of 3D-printed components, mini servos, fishing line, springs, and mechanical fasteners. The completed design included metacarpal bones, phalanges, knuckles, and a wrist mechanism. Finger articulation was achieved through three Emax ES3352 12.4g mini servos per finger: two servos controlled abduction/adduction and base flexion, and a third servo controlled distal flexion, with spring extensions providing passive fingertip retraction (Fig. 1). The total cost of constructing the robotic hand, including all mechanical and electronic components, was approximately \$250, making it a very low-cost platform for research in teleoperation. 

\begin{figure}[htbp]
\centerline{\includegraphics{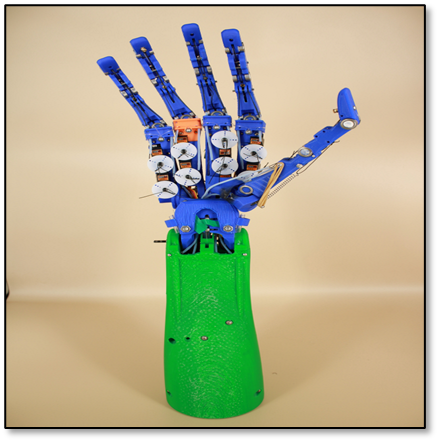}}
\caption{Assembled DexHand (with personal modifications) }
\label{fig}
\end{figure}
\begin{figure*}[htbp]
\centering
\includegraphics[width=\linewidth]{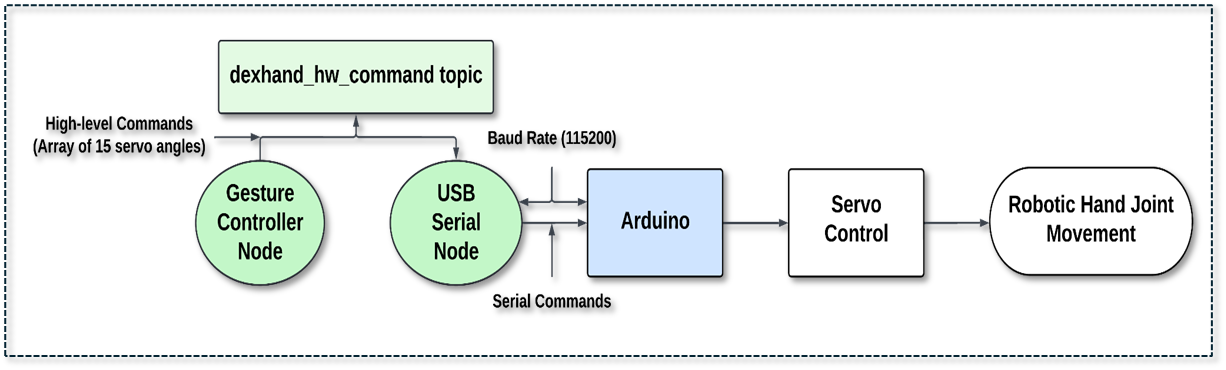}
\caption{ROS 2-Arduino Joint Angle Transmission pipeline for robotic hand servos actuation. }
\label{fig}
\end{figure*}

Control of the robotic hand was implemented using Python via a dual-node ROS 2 system running on a VMware virtual machine with USB passthrough to an Arduino Mega (Fig. 2). A Gesture Controller node generated arrays of 15 servo angles, published to a shared topic (dexhand\_hw\_command), while a USB Serial node formatted and transmitted these commands over serial to the Arduino [10]. The Arduino parsed incoming data using C++ and actuated the corresponding servos. This pipeline enabled smooth, low-latency control of the hand from real-time gesture predictions. 
\subsection{THETA Architectural Pipeline Multi-View Data Collection, Annotation \& Segmentation }\label{AA}
Three RGB webcams were mounted 120 degrees relative to each other and 9 inches in distance away from the center point in order to record the hand from three sides: front, left, and right, giving an accurate 3D view of the hand state (Fig. 3). Every camera used was recording video at a resolution of 640×480 pixels and at a speed of 30 frames per second. Multi-view capture offered more complete hand pose capture, enabling the system to handle occlusion and diversity in hand orientation more effectively.

 \begin{figure}[htbp]
\centerline{\includegraphics{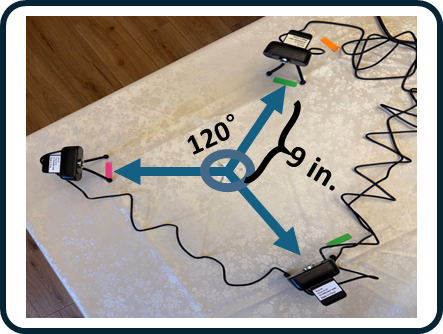}}
\caption{Triangulation Data Collection Setup  }
\label{fig}
\end{figure}
40 different hand postures were selected to record a wide range of finger positions and arrangements. For every gesture, the equivalent set of 15 joint angles was calculated manually with the use of a physical protractor. The joint angles were for the MCP (metacarpophalangeal), PIP (proximal interphalangeal), and DIP (distal interphalangeal) joints of the index, middle, ring, and pinky fingers. 

The measurements of the joint angles were taken carefully and kept in an organized annotation file (gesture\_angles.csv) so that there was an immediate ground truth data for each image frame (Table I). To enhance model robustness and enforce better generalization at inference time, ±5 degrees of jitter was added randomly to all joint angles during data capture. This augmentation simulated realistic variability in gesture appearance across users and viewpoints, preventing the model from overfitting to overly specific static poses.

\begin{table}[htbp]
\caption{Example entries from the “gesture\_angles.csv” dataset}
\centering
\begin{tabular}{|c|c|c|c|}
\hline
\textbf{Gesture ID} & \textbf{Gesture Name} & \textbf{Index MCP Angle} & \textbf{Index PIP Angle} \\
\hline
1 & Closed Fist & 90 ($\pm$5°) & 90 ($\pm$5°) \\
\hline
2 & Open Palm & 180 ($\pm$5°) & 180 ($\pm$5°) \\
\hline
3 & Number One & 180 ($\pm$5°) & 180 ($\pm$5°) \\
\hline
\end{tabular}
\label{tab:gesture_angles}
\end{table}

A custom Python script was developed to automate the sourcing, labeling, and formatting of image and annotation pairs from the dataset for seamless integration into the model training pipeline. The final dataset contained more than 48,000 labeled images, split evenly among all gesture classes and views. The dataset served as the foundation for model training and testing and offered a rich, variegated, and balanced collection of human hand gestures to the model for precise joint angle estimation. 

To isolate the hand before joint angle estimation, we used DeepLabV3, a state-of-the-art semantic segmentation network (Fig. 4). Its architecture employs Atrous Spatial Pyramid Pooling (ASPP) to aggregate multi-scale contextual information, enabling precise segmentation of complex hand structures under diverse poses and lighting conditions.

\begin{figure*}[htbp]
\centering
\includegraphics[width=1\linewidth]{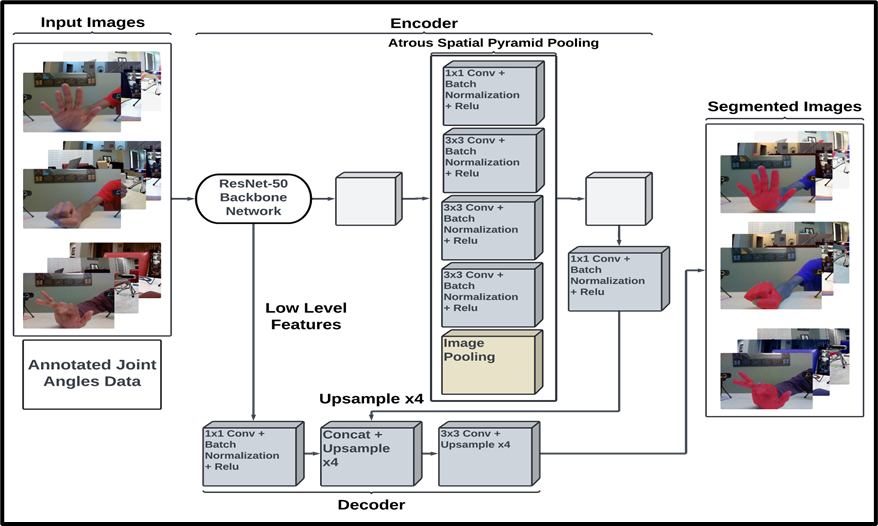}
\caption{Multi-View RGB Image Segmentation Using DeepLabV3 for Image Preprocessing, Feature Extraction, Segmentation Prediction, and Mask Generation. }
\label{fig}
\end{figure*}
The segmentation network uses ResNet-50, a deep residual CNN pretrained on the COCO dataset. Early layers were frozen to preserve low-level feature extraction, while later layers were fine-tuned for binary hand segmentation. ResNet-50 leverages identity-based residual connections to mitigate vanishing gradients and enable efficient optimization of deep architectures. These connections facilitate gradient flow during backpropagation, allowing the network to learn hierarchical features critical for segmenting complex hand shapes and articulations.

\renewcommand{\arraystretch}{1.2} 

\begin{table}[htbp]
\caption{Comparison of Semantic Segmentation Models: mIoU and Parameters}
\centering
\begin{tabular}{|>{\centering\arraybackslash}m{3.5cm}|>{\centering\arraybackslash}m{2cm}|>{\centering\arraybackslash}m{2cm}|}
\hline
\textbf{Model} & \textbf{mIoU (\%)} & \textbf{Parameters (M)} \\
\hline
DeepLab V3-Res50 (2018) & 78.0 / VOC12 & 42 \\
\hline
DeepLab V3+ Xception-71 (2018) & 89.0 / VOC12 & 59 \\
\hline
DeepLab V3+ MobileNetV3 (2021) & 71.0 / Cityscapes & 11 \\
\hline
BiSeNet V1 (2018) & 68.4 / Cityscapes & 5.8 \\
\hline
Fast-SCNN (2019) & 68.0 / Cityscapes & 1.1 \\
\hline
PSPNet R101 (2018) & 78.4 / Cityscapes & 65 \\
\hline
FPN R50 (2018) & 75.9 / Cityscapes & 37 \\
\hline
DenseASPP (2018) & 80.6 / VOC12 & 25 \\
\hline
HRNet-OCR (2020) & 84.5 / Cityscapes & 70 \\
\hline
SegFormer-B5 (2021) & 84.0 / Cityscapes & 85 \\
\hline
\end{tabular}
\label{tab:segmentation_models}
\end{table}
ResNet-50 was selected as the segmentation backbone for its balance between accuracy and complexity (Table II). DeepLab V3-Res50 achieves 78.0\% mIoU on VOC12 with 42 million parameters, offering a strong trade-off between accuracy and computational cost. Lightweight models such as MobileNetV3 (71.0\%, 11M), Fast-SCNN (68.0\%, 1.1M), and BiSeNet V1 (68.4\%, 5.8M) are more efficient but insufficient for precise hand segmentation, while higher-performing models like DeepLab V3+ Xception-71 (89.0\%, 59M), HRNet-OCR (84.5\%, 70M), and SegFormer-B5 (84.0\%, 85M) impose high computational demands. ResNet-50 thus provides effective feature extraction with manageable complexity. The pipeline supports grayscale, RGB, and RGB-D inputs, with the initial layer adapting to input channels. Frames are resized to 224×224, normalized, and processed by a COCO-pretrained DeepLabV3 model using Atrous Spatial Pyramid Pooling. The soft mask output is thresholded at 0.5, refined by morphological operations, and passed to the joint angle estimation module. 

\subsection{THETA Architecture Pipeline Segmentation Preprocessing \& Joint Angle Classification  }\label{AA}
The next step in the THETA pipeline is joint angle prediction from hand images segmented by HSV color thresholds, which isolate red hand regions and remove background pixels. Segmented views from front, right, and left cameras are combined into a multi-view input and processed by a deep learning classifier for joint angle estimation  (Fig. 5). 

\begin{figure*}[htbp]
\centering
\includegraphics[width=\linewidth]{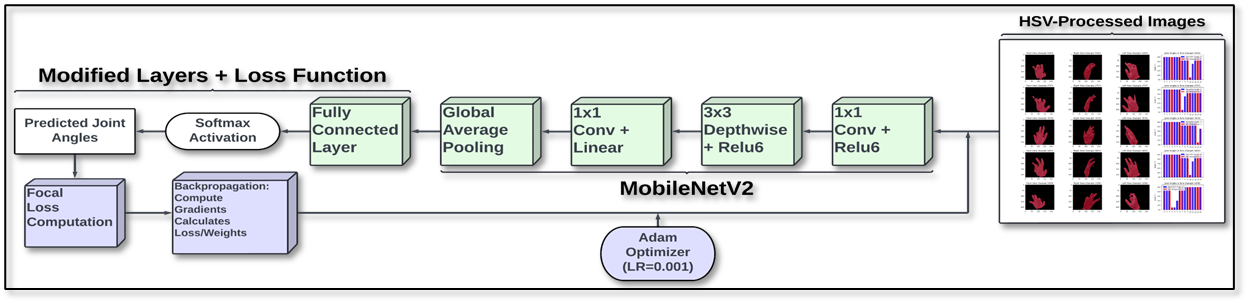}
\caption{Multi-View RGB Image Segmentation Using DeepLabV3 for Image Preprocessing, Feature Extraction, Segmentation Prediction, and Mask Generation. }
\label{fig}
\end{figure*}

\begin{table}[htbp]
\caption{Comparison of Lightweight and Efficient Image Classification Models}
\centering
\begin{tabular}{|>{\centering\arraybackslash}m{3.8cm}|>{\centering\arraybackslash}m{15mm}|>{\centering\arraybackslash}m{2cm}|}
\hline
\textbf{Model} & \textbf{Top-1 Accuracy (\%)} & \textbf{Parameters (M)} \\
\hline
MobileNet V2 (2018) & 72.0 & 3.4 \\
\hline
MobileNet V3-Large (2019) & 75.2 & 5.4 \\
\hline
MnasNet-A1 (2019) & 75.2 & 3.9 \\
\hline
EfficientNet-B0 (2019) & 77.1 & 5.3 \\
\hline
EfficientNet-B7 (2019) & 84.3 & 66 \\
\hline
EfficientNet V2-S (2021) & 83.9 & 22 \\
\hline
EfficientNet V2-L (2021) & 85.7 & 120 \\
\hline
RepVGG-A2 (2021) & 80.2 & 80 \\
\hline
ConvNeXt-Tiny (2022) & 82.5 & 28.6 \\
\hline
CoAtNet-0 (2021) & 81.6 & 25 \\
\hline
\end{tabular}
\label{tab:efficient_models}
\end{table}
This study compared lightweight architectures, including MobileNetV3, EfficientNet, and ConvNeXt (Table III). While MobileNetV3 improves MobileNetV2 in accuracy (75.2\% vs. 72.0\%) with higher complexity (5.4M vs. 3.4M parameters), EfficientNet and ConvNeXt reach up to 85.7\% accuracy but at significantly higher costs (up to 120M parameters), making them less suitable for joint angle prediction. MobileNetV2 provides an optimal trade-off between accuracy, efficiency, and real-time inference for this task. 

MobileNetV2 leverages depthwise separable convolutions and inverted residual blocks to enable rapid feature extraction with low computational overhead, making it well-suited for multi-view hand inputs. The model outputs logits shaped as (batch\_size, 15, 10), corresponding to prediction scores for 15 joints over 10 discrete angle bins. To mitigate class imbalance and overconfidence in dominant classes, softmax activation with temperature scaling (T=2.0) was applied to calibrate the output probability distribution by smoothing predictions, effectively reducing model overconfidence and improving generalization. Additionally, focal loss was combined with inverse bin frequency weighting, assigning higher loss penalties to underrepresented joint poses to address severe class imbalance across the joint angle bins, ensuring balanced learning across common and rare joint configurations. The model was trained for 10 epochs using distributed data parallelism and the Adam optimizer (learning rate 0.001), with transfer learning applied by freezing all but the final two layers. The MobileNetV2-based THETA pipeline demonstrated high accuracy, robustness, and real-time performance in joint angle prediction from segmented multi-view RGB inputs, validating its effectiveness for robotic hand teleoperation. 

 \section{Results \& Analysis }
The model achieved a training accuracy of 97.50\% and a validation accuracy of 97.03\%, with both training and validation losses converging to 0.0001 (Fig. 6). The close alignment between training and validation accuracy, along with minimal final loss, indicates strong generalization and effective mitigation of overfitting. Training and validation losses represent the mean error between predicted and ground-truth joint angles across the respective datasets. 

 \begin{figure}[htbp]
\centerline{\includegraphics[width=1\linewidth]{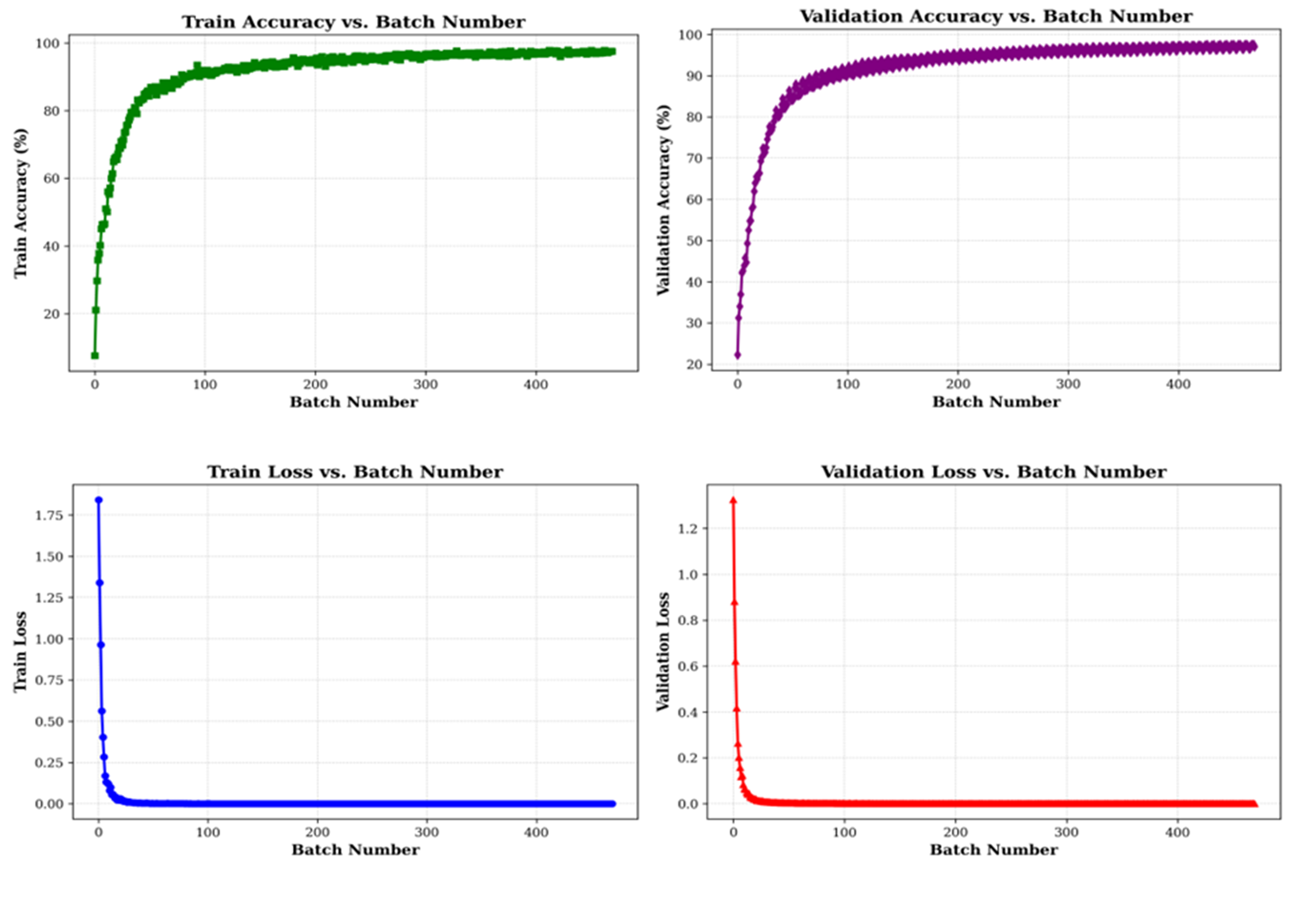}}
\caption{97.50\% training accuracy and 97.03\% validation accuracy with loss convergence to 0.0001.  }
\label{fig}
\end{figure}
THETA was evaluated using accuracy, precision, recall, and F1 score to assess its predictive performance on unseen hand states (Fig. 7). The model achieved 97.18\% testing accuracy, with a precision of 0.8906 and a recall of 0.9872, indicating high correctness and strong detection of true joint angles. The F1 score of 0.9274 confirms a strong balance between precision and recall, particularly important for imbalanced angle distributions. 
 \begin{figure}[htbp]
\centerline{\includegraphics[width=1\linewidth]{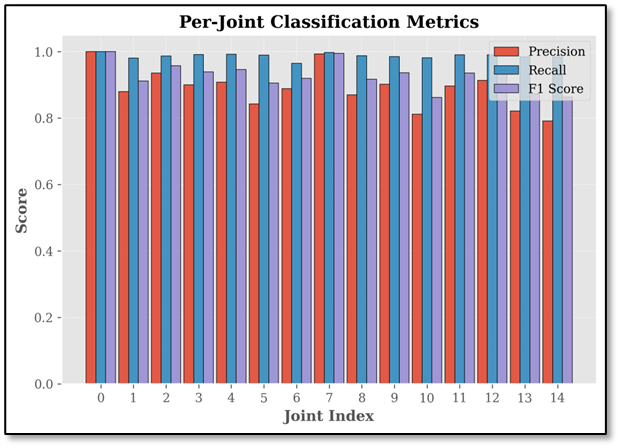}}
\caption{THETA achieves 97.18\% accuracy, 0.9274 F1-score, 0.8906 precision, and 0.9872 recall in joint angle classification, ensuring precise hand pose estimation for robust motion analysis. }
\label{fig}
\end{figure}

The THETA pipeline was validated in real-time using live webcam input and deployed on a servo-actuated robotic hand (Fig. 8). Predicted joint angles showed strong alignment with actual angles across various gestures and lighting conditions. Processed angles were transmitted via serial communication to an Arduino microcontroller as formatted strings, enabling rapid servo actuation. The robotic hand accurately replicated human finger movements, including flexion, extension, and split-finger poses, demonstrating precise, low-latency control through an efficient, low-cost machine learning-based teleoperation framework (Fig. 9). 

 \begin{figure}[htbp]
\centerline{\includegraphics[width=1\linewidth]{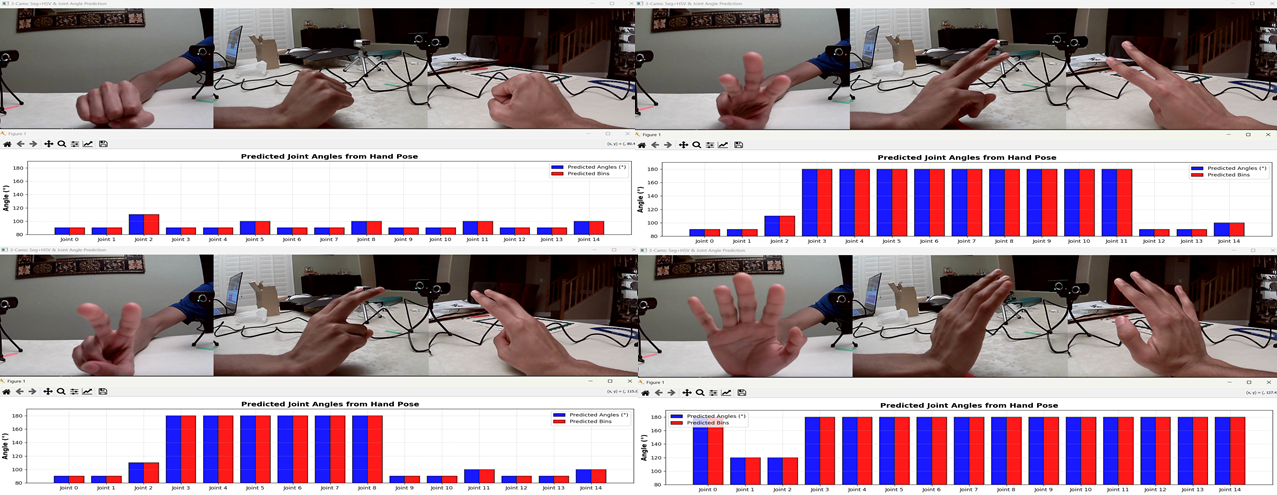}}
\caption{Real-time joint angle inference using THETA’s multi-view triangulation for precise and responsive robotic hand control. }
\label{fig}
\end{figure}
 \begin{figure}[htbp]
\centerline{\includegraphics[width=1\linewidth]{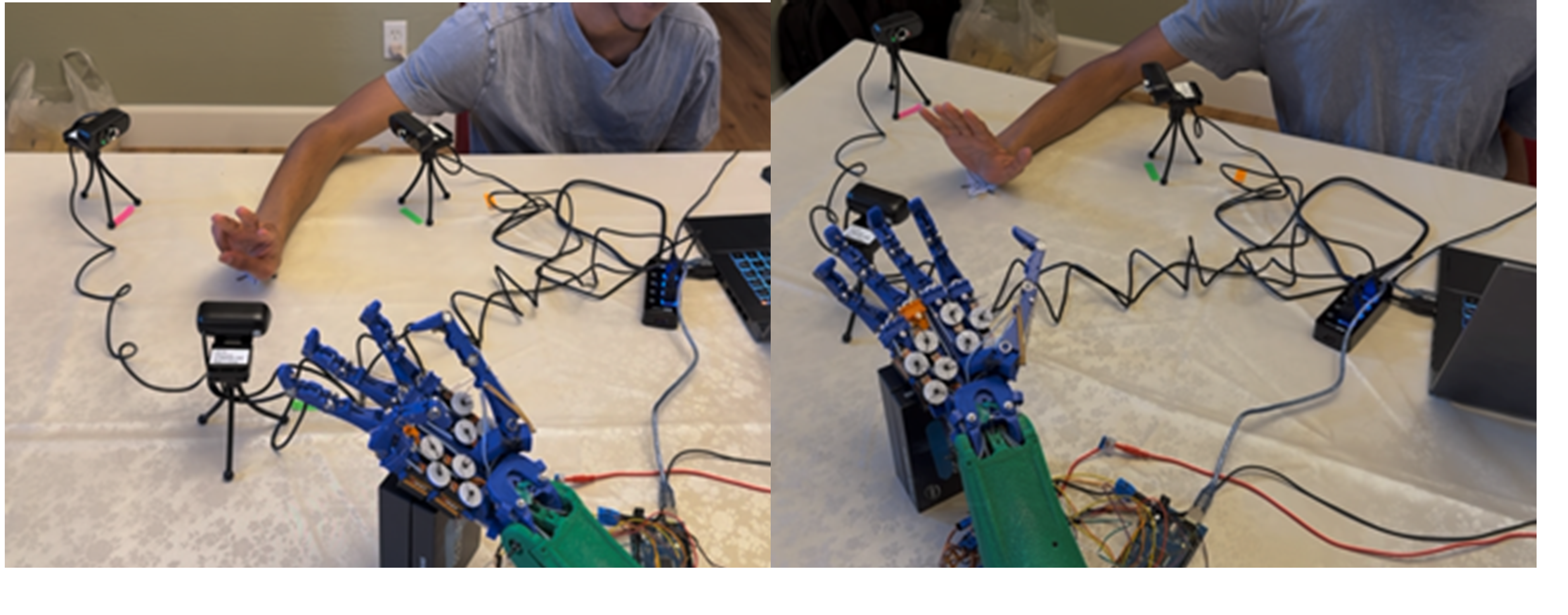}}
\caption{THETA real-time joint angle prediction and inference with serial communication to the DexHand using Arduino. }
\label{fig}
\end{figure}

 \section{Conclusions \& Future Works }
The THETA pipeline provides an efficient deep learning-based framework for real-time robotic hand control, predicting finger joint angles from multi-view RGB inputs with high accuracy, low latency, and minimal hardware requirements. Despite its strong performance, limitations remain. The current dataset, though comprising over 48,000 labeled images, requires further augmentation for better generalization across users, gestures, and environmental conditions. Cloud-based training is computationally expensive, limiting frequent model updates. Additionally, the classification-based architecture discretizes joint angles into bins, reducing smoothness for tasks requiring fine-grained motion.

Future work will address these challenges by shifting to a regression-based model for continuous joint angle prediction and improving articulation precision. Adaptive learning through user-specific feedback will enable personalization over time. Integrating large language models for contextual reasoning may further enhance system performance in dynamic environments. THETA’s compact design, low computational cost, and strong predictive capability make it suitable for applications such as prosthetic control, robotic-assisted surgery, sign language translation, and remote teleoperation in extreme environments.

THETA’s simple setup and robustness has the potential to increase the accessibility of high-compliant teleoperated robotic hands, with implications for countless real-life applications.

\section*{Acknowledgment}

This work was supported by the ASU Sun Robotics Lab, which provided funding, resources, and research support for the development of the THETA pipeline.

\end{document}